\documentstyle[12pt,mkstyle,twoside]{article}
\title{%
On (Anti)Conditional Independence in 
Dempster-Shafer Theory
}

\shorttitle{On (Anti)Conditional Independence in DST}
\date{}
\newcommand{\Bem}[1]{}
\newcommand{\indep}{\bot}
\version{02}

\newcommand{\SP}[1]{\mbox{ \scriptsize
\begin{tabular}{c} #1
\end{tabular}
}}

\newcommand{\eqn}[1]{eqn(\ref{#1})}

\begin{document}
\machetitel

\begin{abstract}
This paper verifies a result of \cite{Shenoy:94} concerning graphoidal
structure of Shenoy's notion of independence for Dempster-Shafer theory of
belief functions.
Shenoy proved that his notion of independence has graphoidal properties
for positive normal valuations. 
 The requirement of strict positive normal valuations as
prerequisite for application of graphoidal properties excludes a wide class of
DS belief functions. It excludes especially so-called probabilistic belief
functions. It is demonstrated that the requirement of positiveness of
valuation may be weakened in that it may be required that commonality
function is non-zero for singleton sets instead, and the graphoidal
properties for independence of belief function variables are then preserved.
This means especially that  probabilistic belief
functions with all singleton sets as focal points  possess  graphoidal
properties for independence  .
\end{abstract}

\section{Introduction}

The concept of conditional independence between two subsets of variables given
a third, well-known from  probability theory \cite{Dawid:79,Pearl:87},  has
also been extensively
studied for other types of uncertainty measures in artificial intelligence,
e.g. for Dempster-Shafer belief function theory 
\cite{Shenoy:94,Studeny:89,Cano:93},
 Spohn's epistemic belief theory \cite{Spohn:88,Hunter:91}, Zadeh's
possibility theory \cite{Shenoy:94,Cano:93}.

The concept of conditional independence in probability theory has been
interpreted in terms of relevance, that is given three disjoint subsets of
variables r,s and t, then r and s are conditionally independent given t means
that the conditional distribution of r given any values of s and t is
governed by the value of t alone; the value of s is irrelevant.
 
The conditional independence relation between subsets of variables in
probability theory possesses many interesting properties
allowing for qualitative reasoning about relevance of sets of variables.
 Pearl and Paz
\cite{Pearl:87} have isolated a subset of these properties called the
"graphoidal
axioms". These axioms are satisfied by several ternary relations beside
probabilistic independence and therefore allow for a wider use of techniques
of qualitative reasoning about relevance for other calculi than probability
calculus. This is especially true of Shenoy's valuation-based system concept
of independence \cite{Shenoy:94} as well as for Cano at al. directed acyclic
graph framework \cite{Cano:93}.

One of important issues closely related to graphoidal structures
is the possibility of 
factorization 
of a joint uncertainty distribution (or, as
called by Shenoy, joint valuation).
Factorization as such may,
for some calculi (e.g. the probability
theory, Dempster-Shafer theory, possibility theory), be used for uncertainty
propagation \cite{Cano:93,Shenoy:90}.
The interesting question is then to what extent  
factorization suitable for qualitative reasoning about relevance (graphoid)
can be used for purposes of uncertainty propagation and vice versa.

We have been interested particularly in factorization of Dempster-Shafer
belief function for purposes of later use in uncertainty propagation. 
We have
shown \cite{Klopotek:93f} that no  factorization 
 may have simpler hypertree structure (required for Shenoy/Shafer's
propagation scheme \cite{Shenoy:90}) than   one made of (in some sense)
conditional factors.  On the other hand, 
 Cano
et al.\cite{Cano:93} and  Shenoy \cite{Shenoy:94} elaborated axiomatic
frameworks within which any factorization of a belief function has graphoidal
properties.
However, our notion of conditionality (called here subsequently
anti-conditionality) and hence of conditional independence differs to some
extent from that of Cano
et al.\cite{Cano:93} and  Shenoy \cite{Shenoy:94}, in that axiomatic
frameworks of
\cite{Cano:93} and  \cite{Shenoy:94} impose more severe restrictions onto the
class of Dempster-Shafer belief functions considered.
As a consequence, there exist belief functions having hypertree factorizations
in general, but not having  equivalent  hypertree  factorizations 
either in Cano
et al. or in Shenoy framework.\\
Hence there exists a gap between the class of factorizations for purposes of
uncertainty propagation as proposed by Shenoy and Shafer \cite{Shenoy:90} and
factorizations  known to have graphoidal properties. 
 The question  emerges
whether or not the classes of DS belief function decompositions fulfilling
graphoidal
axioms can also be widened beyond those considered in \cite{Cano:93} and
 \cite{Shenoy:94}, and especially whether the notion of conditionality and
conditional decomposition as introduced in \cite{Klopotek:93f} is suitable
for this purpose.

The outline of the paper is as follows: in section 2 basic definitions of DST
are recalled. In section 3 the class of belief functions considered by Cano et
al. \cite{Cano:93} is explained. Section 4 presents the class of belief
functions considered by Shenoy \cite{Shenoy:94}. Section 5 presents our
extension to the class of belief functions fulfilling the graphoidal axioms. 
Some consequences are discussed in section 6.\\

\section{Basic Definitions of DST}

Let us first remind basic definitions of DST:
 
\begin{df}
\label{DEFm}
Let $\Xi$ be a finite  set of elements called elementary events. 
Any subset of $\Xi$ be a composite event. $\Xi$ be called also the 
frame of discernment.\\
A basic probability assignment function is any function m:$2^\Xi  \rightarrow
[0, 1]$ such that  $$  \sum_{A \in 2^\Xi } |m(A)|=1, \qquad
  m(\emptyset)=0, \qquad 
\forall_{A \in 2^\Xi} \quad  0 \leq  \sum_{A \subseteq B} m(B)$$
($|.|$ - absolute value)\\
      
A belief function be defined as Bel:$2^\Xi \rightarrow [0,1]$ so that 
 $Bel(A) = \sum_{B \subseteq A} m(B)$
A plausibility function be Pl:$2^\Xi \rightarrow [ 0,1]$  with 
$\forall_{A \in 2^\Xi} \  Pl(A) = 1-Bel(\Xi-A )$
A commonality function be Q:$2^\Xi-\{\emptyset\} \rightarrow [0,1]$ with 
 $\forall_{A \in 2^\Xi-\{\emptyset\}} \quad Q(A) = \sum_{A \subseteq B}
m(B)$ \end{df}

If for every $A\subseteq\Xi$ we have $m(A)\ge 0$, then we talk about proper
belief functions. 
If for every $A\subseteq\Xi$ we have $Q(A)\ge 0$, then we talk about pseudo-%
belief functions. 

Furthermore, a Rule of Combination of two Independent Belief Functions 
$Bel_1$,
 $Bel_2$ Over the Same Frame of Discernment (the so-called Dempster-Rule),
denoted 
    $$Bel_{E_1,E_2}=Bel_{E_1} \oplus Bel_{E_2}$$ 
 is defined as follows: :
$$m_{E_1,E_2}(A)=c \cdot  \sum_{B,C; A= B \cap C} m_{E_1}(B) \cdot 
m_{E_2}(C)$$ (c - constant normalizing the sum of $|m|$ to 1)

Furthermore, let the frame of discernment $\Xi$ be structured in that it is
identical to cross product of domains $\Xi_1$, $\Xi_2$, \dots $\Xi_n$ of n
discrete variables $X_1, X_2, \dots X_n$, which span the space $\Xi$. Let
$(x_1, x_2, \dots x_n)$ be a vector in the space spanned by the variables 
$X_1,
 ,  X_2, \dots X_n$. Its projection onto the subspace spanned by variables 
$X_{j_1}, X_{j_2}, \dots X_{j_k}$ ($j_1, j_2,\dots j_k$ distinct indices from
the set 1,2,\dots,n) is then the vector $(x_{j_1}, x_{j_2}, \dots x_{j_k})$. 
$(x_1, x_2, \dots x_n)$ is also called an extension of $(x_{j_1}, x_{j_2},
\dots x_{j_k})$. A projection of a set $A$ of such vectors is the set
$A ^{\downarrow \{X_{j_1}, X_{j_2}, \dots X_{j_k}\}}$ 
 of
projections of all individual vectors from A onto $X_{j_1}, X_{j_2}, \dots
X_{j_k}$. A is also called an extension of $A ^{\downarrow \{X_{j_1}, X_{j_2},
\dots X_{j_k}\}}$. A is called the vacuous extension of $A ^{\downarrow
\{X_{j_1},
 X_{j_2}, \dots X_{j_k}\}}$  iff A contains all possible extensions of each
individual vector in $A ^{\downarrow \{X_{j_1}, X_{j_2}, \dots X_{j_k}\}}$ .
The fact, that A is a vacuous extension of B onto space $X_1,X_2,\dots\,
X_n$ is denoted by $A=B ^{\uparrow \{X_1,X_2,\dots\,X_n\}}$
\begin{df} (see \cite{Shenoy:90})
Let m be a basic probability assignment function on the space of discernment
spanned by variables   $X_1,X_2,\dots\,X_n$. $m ^{\downarrow \{X_{j_1},
X_{j_2}, \dots X_{j_k}\}}$ is  called  the  projection  of  m  onto 
subspace spanned by
$X_{j_1}, X_{j_2}, \dots X_{j_k}$ iff 
$$m ^{\downarrow \{X_{j_1}, X_{j_2}, \dots X_{j_k}\}}(B)= c \cdot
\sum_{A; B=A  ^{\downarrow \{X_{j_1}, X_{j_2}, \dots X_{j_k}\}} } m(A)$$
(c - normalizing factor)
\end{df}
\begin{df}  (see \cite{Shenoy:90})
Let m be a basic probability assignment function on the space of discernment
spanned by variables  $  X_{j_1},
X_{j_2}, \dots X_{j_k} $. $m ^{\uparrow \{X_1,X_2,\dots\,X_n\}}$ is called
the vacuous extension 
 of m onto superspace spanned by $X_1,X_2,\dots\,X_n$
iff 
$$m ^{\uparrow \{X_1, X_2, \dots X_n\}}
(B ^{\uparrow \{X_1,X_2,\dots\,X_n\}})=m(B)$$
\noindent
and $m ^{\uparrow \{X_1, X_2, \dots X_n\}}(A)=0$ for any other A. \\
We say that a belief function is vacuous iff $m(\Xi)=1$ and $m(A)=0$ for any A
different from $\Xi$.
\end{df}

Projections and vacuous extensions of Bel, Pl and Q functions are defined
with
respect to operations on m function. Notice that, by convention, if we want to
combine by Dempster rule two belief functions not sharing the frame of
discernment, we look for the closest common vacuous extension of their
frames of discernment without explicitly notifying it.

\begin{df} \label{DEFcond}
 (See \cite{Shafer:90b}) Let B be a subset of $\Xi$, called 
evidence,
 $m_B$ be a basic probability assignment such that $m_B(B)=1$ and $m_B(A)=0$
for any A different from B. Then the conditional belief function $Bel(.||B)$
representing the belief function $Bel$ conditioned on evidence  B 
is defined
as: $Bel(.||B)=Bel \oplus Bel_B$. 
\end{df}
\begin{df} \label{DEFuncindep} (See \cite{Shenoy:94}
Two disjoint sets of variables $p.q$ are said to be (unconditionally)
independent) iff 
$$Bel ^{\downarrow p\cup q}=
Bel ^{\downarrow p}\oplus
Bel ^{\downarrow  q}$$
\end{df}

\section{Cano's et al. A Priori Conditionals in Directed Acyclic Graphs}

Cano et al. in \cite{Cano:93} proposed a generalization of Pearl's bayesian
networks \cite{Pearl:88} to represent DS belief distribution factorization. 
They motivated their choice by stating that "graphical
structures used to represent
relationships among variables in our      work are     Pearl's causal networks
\cite{Pearl:88}, not Shenoy/Shafer's hypergraphs \cite{Shenoy:90}, because the
former are
more appropriate to represent independence relationships among variables in a
direct way." (p.257).
They discovered also that Dempster-Shafer theory needs two types of
conditionality - the one introduced by Shafer \cite{Shafer:90b}
(see definition \ref{DEFcond} above)
 which they
call a-posteriori conditionality, which is not suitable for generalization of
bayesian belief networks, and a different one, which they call a-priori
conditionality. On page 262 (Definition 2) they define
a belief function $Bel$ to be
 (a priori)
conditional belief function conditioned on variable set  $h$ 
by  requiring  $Bel ^{\downarrow h}$ to be a vacuous belief function. 
This latter notion clearly generalizes probabilistic conditionality in a way
allowing for usage of probabilistic algorithms for uncertainty propagation. 
However, it cannot handle various cases of functions which could
be factored in terms of a Dempster Rule of Combination.

As an example please verify, that the belief function $Bel_{12}$
$$Bel_{12}=Bel_1\oplus Bel_2$$

\noindent
with focal points for $Bel_1$, $Bel_2$ ($Bel_1$ defined for variables X,Y,
$Bel_2$ for variables X,Z, domains of variables: X: \{$x_1,x_2$\},  
Y: \{$y_1,y_2$\},  
Z: \{$z_1,z_2$\})
  
\begin{center}
\begin{tabular}{ll}
\multicolumn{2}{c}{$Bel_1$}\\
set & $m_1(set)$\\
\hline
\{$(x_1,y_1), (x_1,y_2),$\\ $ (x_2,y_1), (x_2,y_2)$\}        &    0.1\\
\{$(x_1,y_1)$\}        &    0.2\\
\{$(x_1,y_2)$\}        &    0.25\\
\{$(x_2,y_1)$\}        &    0.3\\
\{$(x_2,y_2)$\}        &    0.15\\
\hline \quad\\
\end{tabular} 
\begin{tabular}{ll}
\multicolumn{2}{c}{$Bel_1$}\\
set & $m_2(set)$\\
\hline
\{$(x_1,z_1), (x_1,z_2),$\\$ (x_2,z_1), (x_2,z_2)$\}        &    0.2\\
\{$(x_1,z_1)$\}        &    0.2\\
\{$(x_1,z_2)$\}        &    0.3 \\
\{$(x_2,z_1)$\}        &    0.25\\
\{$(x_2,z_2)$\}        &    0.05\\
\hline \quad\\
\end{tabular} 
\end{center}

\noindent
cannot be represented in a structured manner as a product  of  an 
unconditional
 and (a priori) conditional belief function in sense of Cano et al.

\section{Shenoy's Notion of Conditionality}

Shenoy \cite{Shenoy:94} introduced a totally different
notion of conditionality within his
framework of Valuation-Based Systems (VBS). 
Its starting point is to define conditional independence in terms of 
factorization.
VBS is defined by a system of
axioms presented in that paper. 
Basic concepts of VBS are: 
the notion of a set of variables s introduced on page 205, 
the notion of set of valuations $V_s$ introduced on page 206,
the
notion of set $N_s$ of normal valuations on page 207, the notion of the set
$U_s$ of positive normal valuations (that of those normal valuations which
have
unique identity with respect to the valuation 
combination operator $\oplus$) introduced on page
208-209.
He defined conditional independence as follows:
 (Definition 3.1. p. 214)
Suppose $\gamma \in N_w$, suppose r,s,v are disjoint subsets of (the set of
variables) w. Let $\gamma(t)$, $t\subseteq w$ denote projection of $\gamma$
onto subspace spanned by variables t. We say that
r and s are conditionally independent given v with respect to  $\gamma$,
written as $r \indep_\gamma s|v$ iff there exist 
$\alpha_{r \cup v} \in V_{r \cup v}$  and 
$\alpha_{s \cup v} \in V_{s \cup v}$  such that
$$\gamma(r\cup s\cup v)= 
\alpha_{r \cup v} \oplus
\alpha_{s \cup v} $$

To prove that his notion of conditional independence 
(Definition 3.1.) is a graphoid (a concept defined in \cite{Pearl:87}),
he
proves graphoidal properties in theorems 3.1.(symmetry), 3.2. (decomposition),
3.3.(weak union), 3.4.(contraction) and 3.5. (intersection).  
Theorems 3.1.-3.4 are valid for normal valuations.
 The notion of positive
normal valuation is used in theorem 3.5 (intersection) p.219.
  The proof of theorem 3.5 relies on the particular form
of claim (7) of Lemma 3.1., and on the fact that individual valuation identity
turns to group identity in $N_s$ if positiveness is assumed.

Let us cite Shenoy's 
Lemma 3.1., (page 215)
claim 7, because it will constitute the central point of our further interest
:  $r \indep s | v$ equivalent to
 $\gamma(r|s \cup v)=\gamma(r|v) \oplus \tau_{\gamma(s\cup v)}$

The notion of conditionality ($\gamma(r|v)$)  is introduced on 
page 213. " Suppose $\sigma \in N_s$ and suppose a and b are disjoint subsets
of s.. .... Let $\sigma(b|a)$ denote $\sigma
^{\downarrow a \cup b} \ominus \sigma ^{\downarrow b}$." 
 The removal operator $\ominus$ has been described by axioms R1,R2 and CR on
page 212. 
The $\tau_{\sigma}$ - the member identity - has been defined
in axiom R2 on page 212. 

We will omit here the citation of  general definitions of the above-mentioned
terms, as they are lengthy and complicated, 
 but
we will concentrate on their meaning for the Dempster-Shafer theory, as it is
our main point of interest.

On page 224 the above-mentioned  notions are specialized for DST:

 A {\em valuation}
for s is a function $\sigma:2^{W_s}\leftarrow[0,1]$. This function $\sigma$
is the commonality function Q of DST. (We prefer notation $\Xi_s$ in place of
$W_s$ for universe spanned by variable set $s$; but we keep in this section
denotation of Shenoy for easier cross-reference). 

 $\sigma$ is {\em normal} iff
$\Sigma_{a \in 2^{W_s}}  ((-1)^{|a|+1}\sigma(a) =1$. This means actually that
the sum of all masses over all focal points has  to  be  equal  1 
(this differs a
bit from definition
\ref{DEFm} in this paper, as we assumed that the sum of absolute values of
the mass function over all focal points has to be equal 1. This results in a
difference in scaling factor, but has no further effect). 

  On page 225 the {\em removal}
operator is introduced for DST. Suppose $\sigma \in V_s$ and $\rho \in N_s$.
Let $K=\Sigma_{a \in 2^{W_s}, \rho(a )>0) }
((-1)^{|a|+1}\sigma(a )/ \rho(a) ) $. Then 
if $K>0$ and $\rho(a )>0$ then $(\sigma\ominus\rho)(a)=K ^{-1}
\sigma(a)/\rho(a)$ and otherwise $ (\sigma\ominus\rho)(a)=0$.\\
This means that the removal operator is defined
for every set 
 as division of commonality
functions whenever the second commonality function takes positive  values and
as 0 elsewhere, and the division is followed by normalization of mass
function.

This implies that {\em conditioning } on the set of variables  $v$ in DST in
Shenoy's framework is defined as  division of commonality
function
by its projection onto the set of variables  $v$
 whenever the projected Q-function  takes positive  values and
as 0 elsewhere, and the division is followed by normalization of mass
function of the result. 

Under these circumstances the {\em group identity } for the space of normal
valuations is a belief function with the only focal point $m(W_s)=1$, where
$W_s$ is the universe (spanned by variables from set $s$). Member identities
are usually
complex constructs with masses taking (positive and negative) integer 
values.

Obviously, then a valuation 
 $\sigma$  is {\em positive normal } iff $\sigma(a)>0$ for every $a\in
2^{W_s}$. This means that  $m(W_s)>0$, where
$W_s$ is the universe.

 Notions of
conditionality of Shenoy \cite{Shenoy:94} and of Cano et al. \cite{Cano:93}
are different in general. But regrettably, in the interesting case of
graphoidal
properties positive normality is required. And  only for positive normal
valuations in  Dempster-Shafer theory
 notions of
conditionality of Shenoy \cite{Shenoy:94} and of Cano et al. \cite{Cano:93}
coincide (clearly in case when Cano conditionals exist at all) ! This actually
means the following:
\begin{itemize}
\item Three exist belief functions which possess graphoidal decompositions in
sense of Cano et al. and in the sense of Shenoy such that qualitative
relevance results agree
\item Three exist belief functions which possess graphoidal decompositions 
in the sense of Shenoy such that qualitative
independence between sets of variables p,q given r is granted in Shenoy's
decomposition but no such decomposition in the sense of Cano et. al exists.
This especially true for the example given at the end of previous section. 
\item Three exist belief functions which possess graphoidal decompositions 
in the sense of Cano et al. such that qualitative
independence between sets of variables p,q given r is granted in Cano's
decomposition but no such decomposition in the sense of Shenoy   exists.
It is the case for probability distributions. Probability distributions are
considered as a special case of DS belief functions with focal points only on
singleton sets. Hence they are not positive valuations in the sense of Shenoy
(because probabilistic belief functions have more than one valuation
identity). )
\end{itemize}


Last not least let us notice that the notion of conditionality
 leads outside
the domain of proper belief functions of DST (those with nonnegative mass
function) and shifts the considerations into the area of pseudo-belief
(those with non-negative commonality function) \cite{Shenoy:94}. 
Shenoy states on  pp.225-226
"Notice that if $\sigma$ and $\rho$ are commonality functions, it is possible
that  $\sigma\ominus\rho$  may  not  be  a  commonality  function 
because condition
... [of non-negativity of mass function] may not be satisfied by
$\sigma\ominus\rho$ In fact, if $\sigma$ is a commonality function for s, and
$r\subseteq s$, then even $\sigma\ominus{\sigma}^{\downarrow r}$ may fail to
be a commonality function. This fact is the reason why we need the concept of
proper valuation as distinct from non-zero and normal valuations in the
general VBS framework. An implication of this fact is that conditionals may
lack semantic coherence 
in the Dempster-Shafer's theory. This is the primary reason why conditionals
are neither natural nor widely studied in the Dempster-Shafer's
belief-function theory".
What is more - as Studeny claims at the end of his paper \cite{Studeny:89} -
even the notion of Shenoy's conditional independence  leads outside the
domain of proper belief functions. that is if $p,q$ are independent given $r$
with respect to proper belief function $Bel$
in the sense of Shenoy (as cited above), then there may NOT
exist two proper belief functions $Bel_1,Bel_2$ such that
$Bel_1$ is defined over space spanned by variables $p\cup r$ and 
$Bel_2$ is defined over space spanned by variables $q\cup r$ and 
$$Bel ^{\downarrow p\cup q \cup r}=  Bel_1 \oplus  Bel_2 $$
holds. 

\section{Main Result}


 Below it is
demonstrated that 
Shenoy's valuation
positiveness is not required in order to achieve truth of
intersection, and this due to the possibility of verifying the contents of
claim (7) of Lemma 3.1. of \cite{Shenoy:94}.

At the very beginning let us clarify why we (as well as other authors) do not
use Shafer's definition of conditionality cited in definition \ref{DEFcond}
when talking about independence. In general, independence is understood in
terms of irrelevance. For example, if in a probability distribution $P$ in
variables $X,Y$ these variables $X,Y$  are mutually independent ($P(Y|X=x_i)$
is the same whatever value $x_i$ of $X$ is considered), then $P(X,
Y)=P(X)cdot P(Y)$ that is the interrelationship of X and Y is irrelevant for
representation
the joint probability distribution. 

But let us take the following belief distribution in variables X,Y, both
variables with domains of cardinality 2.
\begin{center}
\begin{tabular}{l|r}
Focal &  $m$(focal)\\
\hline
\{
$(x_1,y_1)$, 
$(x_1,y_2)$, 
$(x_2,y_1)$
\} & 0.25\\
\hline
\{
$(x_1,y_1)$, 
$(x_1,y_2)$, 
$(x_2,y_2)$
\} & 0.25\\
\hline
\{
$(x_1,y_1)$, 
$(x_2,y_1)$, 
$(x_2,y_2)$
\} & 0.25\\
\hline
\{
$(x_1,y_2)$, 
$(x_2,y_1)$, 
$(x_2,y_2)$
\} & 0.25\\
\hline
\end{tabular}
\end{center}
\noindent
Let $\Xi_Y=\{y_1,y_2\}$. 
It is an easy task to check that for every (non-empty) subset $S$ of the
domain of X $Bel(||S\times\Xi_Y)^{\downarrow Y}$ is the same that is the
marginal distribution in variable Y does not depend on X. But nonetheless $Bel
  ^{\downarrow \{X,Y\}} \neq  Bel
  ^{\downarrow \{X\}} \oplus Bel
  ^{\downarrow \{Y\}}$ as definition \ref{DEFuncindep} would require.

\begin{df}  \label{MKd1}
For belief (or pseudo-belief) function $Bel$ over 
discourse space spanned by the set of variables $V=\{X_1, X_2,....
 X_n\}$ we define (anti)conditional belief function 
$Bel ^{V | p}(A)$ of $Bel$ conditioned on set of variables $p, p\subseteq V$
from the set $V$ as any pseudo-belief function
 fulfilling the equation
\begin{equation}  \label{MKe1}
 Bel=Bel ^{\downarrow p  } \oplus  Bel ^{| p  }
\end{equation}
\end{df}

REMARK: Obviously $Q ^{|p}(A)= c\cdot \frac{Q(A)}{Q ^{\downarrow p  }(A)}$
($c$ - a mass assignment normalizing constant, independent of A)
for every set $A$ such that $Q ^{\downarrow p  }(A) \neq 0 $

\begin{df}
For belief (or pseudo-belief) function $Bel$ over 
discourse space spanned by the set of variables $V=\{X_1, X_2,....
 X_n\}$ we say that $Bel$ is compressibly independent of a set of variables
$p$ from the set $V$ iff      the following equation holds
\begin{equation} \label{MKe2}
 Bel=(Bel ^{\downarrow V-p})^{\uparrow p}
\end{equation}
(that is $Bel$ is in fact a vacuous extension of another belief or
pseudo-belief function defined over space of discourse spanned by the
set of variables $V-p$).
\end{df}

Notice that if
 the belief
function $Bel$ is compressibly independent of the set of variables $p$
then it
can be represented in a "compressed" way by the function
$Bel ^{\downarrow V-p}$.

REMARK: We assume that operators $\downarrow, \uparrow, |$ are of same
priority and are processed from left to right, so that 
e.g. $((Bel ^{\downarrow p})^{|q})^{\uparrow r}$ is equivalent to saying 
 $Bel ^{\downarrow p |q \uparrow r}$. 

Please notice that 
if $Bel$ is a belief function   over 
discourse space spanned by the set of variables $V$ then 
a conditional belief function $Bel ^{|p}$ ($p\subseteq V$) may be
compressibly independent of the set of variables $q$ ($p\cap q=\emptyset,
q\subseteq V$) while at the same time $Bel$ itself may not be compressibly
independent of the set of variables $q$. Consider for example the belief
function Bel in variables X,Y,Z. with domains $\{x_1,x_2,x_3\}$,   $\{y_1,y_2,
y_3\}$,   $\{z_1,z_2, z_3\}$,  and the following single focal point:

\begin{center}
\begin{tabular}{l|r}
Focal &  $m$(focal)\\
\hline
\{
$(x_1,y_1,z_1)$, 
$(x_2,y_2,z_2)$, 
$(x_3,y_3,z_3)$
\} & 1
\end{tabular}
\end{center}
\noindent
so that Bel is neither compressibly independent of X, nor of Y nor of
Z. Furthermore, no partition of the set of variables into independent subsets
in the sense of Shafer is possible, that is 
$Bel \neq Bel ^{\downarrow \{X,Y\}} \oplus Bel ^{\downarrow \{Z\}}$ 
 etc.
Then    $Bel ^{\downarrow \{Y,Z\}}$ has focal point:
\begin{center}
\begin{tabular}{l|r}
Focal &  $m ^{\downarrow \{Y,Z\}}$(focal)\\
\hline
\{
$(y_1,z_1)$, 
$(y_2,z_2)$, 
$(y_3,z_3)$
\} & 1
\end{tabular}
\end{center}
\noindent
and  $Bel ^{\downarrow \{X,Y\}}$ has focal point:
\begin{center}
\begin{tabular}{l|r}
Focal &  $m ^{\downarrow \{X,Y\}}$(focal)\\
\hline
\{
$(x_1,y_1)$, 
$(x_2,y_2)$, 
$(x_3,y_3)$
\} & 1
\end{tabular}
\end{center}
\noindent 
and   $Bel ^{\downarrow \{Z\}}$ has the focal point:
\begin{center}
\begin{tabular}{l|r}
Focal &  $m ^{\downarrow \{Z\}}$(focal)\\
\hline
\{
$(z_1)$, 
$(z_2)$, 
$(z_3)$
\} & 1
\end{tabular}
\end{center}
However, 
an  (anti)conditional belief function $Bel ^{|\{Y,Z\}}$ with
following focal point:
\begin{center}
\begin{tabular}{l|r}
Focal &  $m ^{|\{Y,Z\}}$(focal)\\
\hline
\{
$(x_1,y_1,z_1)$, 
$(x_1,y_1,z_2)$, 
$(x_1,y_1,z_3)$
,  & \\
$(x_2,y_2,z_1)$, 
$(x_2,y_2,z_2)$, 
$(x_2,y_2,z_3)$
,  & \\
$(x_3,y_3,z_1)$, 
$(x_3,y_3,z_2)$, 
$(x_3,y_3,z_3)$
\} & 1
\end{tabular}
\end{center}
\noindent
 is compressibly independent of Z, that is there exists a(n anti)-
conditional $(Bel  ^{\downarrow \{X,Y\}})^{ | \{Y\} } $
\begin{center}
\begin{tabular}{l|r}
Focal &  $m ^{\downarrow \{X,Y\}|\{Y\}}$(focal)\\
\hline
\{
$(x_1,y_1)$, 
$(x_2,y_2)$, 
$(x_3,y_3)$
\} & 1
\end{tabular}
\end{center}
\noindent such
that 
$Bel ^{|\{Y,Z\}}=((Bel  ^{\downarrow \{X,Y\}})^{ | \{Y\} } )
  ^{\uparrow \{X,Y,Z\} }$ 

Please pay attention to the fact that by definition there may be several
distinct (anti)conditional belief functions for a given function
(unless we have to do with positive normal valuations as defined by Shenoy).
Consider for example the belief function in two variables, X,Y with focal
points:

\begin{center}
\begin{tabular}{l|r}
Focal &  $m $(focal)\\
\hline
\{
$(x_1,y_1)$, 
$(x_2,y_2)$, 
\} & 0.75\\
\hline
\{
$(x_1,y_2)$, 
$(x_3,y_3)$
\} & 0.25\\
\hline 
\end{tabular}
\end{center}
\noindent
for which at least two (anti)conditional belief functions $Bel ^{|Y}$ are
possible, one with 
\begin{center}
\begin{tabular}{l|r}
Focal &  $m ^{|\{Y\}}$(focal)\\
\hline
\{
$(x_1,y_1)$, 
$(x_1,y_2)$, 
$(x_2,y_2)$, 
$(x_3,y_3)$
\} & 1\\
\hline 
\end{tabular}
\end{center}
\noindent
and the other with 
\begin{center}
\begin{tabular}{l|r}
Focal &  $m ^{|\{Y\}}$(focal)\\
\hline
\{
$(x_1,y_1)$, 
$(x_2,y_2)$, 
\} & 0.5 \\
\hline
\{
$(x_1,y_2)$, 
$(x_3,y_3)$
\} & 0.5 \\
\hline
\end{tabular}
\end{center}

This differs from
the approach of Shenoy where the conditional belief  had to be unique, and
the 
definition \ref{MKd1} covers both the concept of conditionality of Cano el
al. and of Shenoy. 
(By the way: the first conditional of the two above is in sense of Shenoy,
while the second in the sense of Cano et. al.).
 
It is obvious why ambiguity in definition of conditionals has been avoided by
Shenoy - because
in his lemmas and theorems equations would have to be replaced by existential
statements. Subsequently we demonstrate that the ambiguity of the 
definition  \ref{MKd1} can be handled quite conveniently.\\

\begin{th} \label{MKt1}
Let $p,q,r$ be pairwise disjoint sets of variables and let 
 $V =p \cup  q \cup r $ and let $Bel$ be defined 
 over V.
Furthermore let $Bel ^{|p \cup r}$ be a (anti)conditional Belief
conditioned on variables from  $p\cup r$. Let this conditional distribution be
 compressibly 
 independent of 
$r$. Let $Bel ^{\downarrow p \cup q}$ be the projection of $Bel$ onto 
the subspace spanned by $p\cup q$. Then there exists 
 $Bel ^{\downarrow p \cup q | p}$  being a conditional belief of that 
projected belief conditioned on the variable $p$ such that this  $Bel 
^{|p \cup
r}$  is the vacuous extension of  $Bel ^{\downarrow p \cup q | p}$ 
\begin{equation} \label{MKe3}
Bel ^{|p \cup r} =
  (Bel ^{\downarrow p \cup q | p}) ^{\uparrow V}
\end{equation} 
\end{th}

\AnfBeweis
By definition (see \eqn{MKe1}):

\begin{equation}
Bel=Bel   ^{\downarrow   p   \cup r}
 \oplus  Bel  ^{V | p \cup r}
\end{equation}

and hence (def.\ref{DEFm})
\begin{equation}
m(A)=\sum_{
\SP{
$B,C;$\\
$B,C \subseteq \Xi$,\\
$A=B \cap C$
}
} 
m   ^{\downarrow   p   \cup
r \uparrow V}( C ) \cdot  m ^{|p \cup r} (B)  
\end{equation}

As  we assume the compressible 
 independence of the conditional Belief  $Bel ^{|p \cup
r}$   from the variable set
$r$, so  $m ^{|p \cup r}$ 
is being a vacuous extension of another distribution, say m', 
defined only over $p \cup q$, so we in fact 
calculate the right-hand-side sum as:

\begin{equation} \label{MKe6}
m(A)=\sum_{
\SP{
$b,c; $\\
$b \subseteq \Xi_p \times \Xi_q,$\\
$ c \subseteq  \Xi_p \times \Xi_r,$\\
$  A = b ^{\uparrow V} \cap
  c ^{\uparrow V}$
}
}  
m   ^{\downarrow   p   \cup
r}( c ) \cdot   m'(b)
\end{equation}

Let us marginalize both sides of \eqn{MKe6} over $r$ ($a \subseteq \Xi_p
\times \Xi_q$): 
\begin{equation}
m ^{\downarrow p \cup q}(a) = 
\sum_{A; a = A  ^{\downarrow p \cup q}}
\left(
\sum_{
\SP{
$b,c;$\\
$ b \subseteq \Xi_p \times \Xi_q,$\\
$ c \subseteq  \Xi_p \times \Xi_r,$\\
$  A = b ^{\uparrow V} \cap
  c ^{\uparrow V} $
} 
} 
   m'(b) \cdot  m   ^{\downarrow   p   \cup r}( c )
\right)
\end{equation}

Hence eliminating auxiliary set A we obtain:

\begin{equation} \label{MKe8}
m ^{\downarrow p \cup q}(a) = 
\sum_{
\SP{
$b,c$;\\
$ b \subseteq \Xi_p \times \Xi_q,$\\
$c \subseteq  \Xi_p \times \Xi_r,$ \\
$ a = ( b ^{\uparrow V} \cap c ^{\uparrow V})  
^{\downarrow p \cup q}$\\
}
}  
   m'(b) \cdot  m   ^{\downarrow   p   \cup r}( c )
\end{equation}

It is easily checked that if $b \subseteq \Xi_p \times \Xi_q$ and
 $c \subseteq \Xi_p \times \Xi_r$ 
then 
\begin{equation}  \label{MKe9}
 ( b ^{\uparrow V} \cap c ^{\uparrow V})  
^{\downarrow p \cup q}
= b \cap (c ^{\uparrow V}) 
^{\downarrow p \cup q} 
\end{equation}

But as $c$ is defined over $p \cup r$:\\
\begin{equation}  \label{MKe10}
(c ^{\uparrow V}) ^{\downarrow p \cup q}
= (c  ^{\downarrow p})  ^{\uparrow p \cup q}
\end{equation}

Hence, by substituting \eqn{MKe9} and  \eqn{MKe10} into \eqn{MKe8} we get:
\begin{equation} \label{MKe11}
m ^{\downarrow p \cup q}(a) = 
\sum_{
\SP{
$b,\gamma$;\\
$ b \subseteq \Xi_p \times \Xi_q,$\\
$\gamma \subseteq  \Xi_p, $\\
$a =  b  \cap \gamma ^{\uparrow p \cup q}  $
}
} 
\sum_{
\SP{
$c;$\\
$c \subseteq  \Xi_p \times \Xi_r,$\\
$ \gamma = c ^{\downarrow p}$
}
}
 m'(b) \cdot  m   ^{\downarrow   p   \cup r}( c )
\end{equation}

As $b$ does not depend on $c$ in the inner sum of \eqn{MKe11}, we get 
:
\begin{equation} \label{MKe12}
m ^{\downarrow p \cup q}(a) = 
\sum_{
\SP{
$b,\gamma ;$ \\
$b \subseteq \Xi_p \times \Xi_q,$\\
$\gamma  \subseteq  \Xi_p, $\\
$a =  b  \cap \gamma  ^{\uparrow p \cup q}  $\\
}
}
 m'(b) \cdot  
\sum_{
\SP{
$c;$\\
$c \subseteq  \Xi_p \times \Xi_r,$\\
$ \gamma  = c ^{\downarrow p}$
}
} 
 m   ^{\downarrow   p   \cup r}( c )
\end{equation}

But by definition (of projection in DST) for $\gamma \subseteq \Xi_p$
\begin{equation} \label{MKe13}
\sum_{
\SP{
$c;$\\
$c \subseteq  \Xi_p \times \Xi_r, $\\
$\gamma  = c ^{\downarrow p}$
}
}
 m   ^{\downarrow   p   \cup r}( c ) =m   ^{\downarrow p}( \gamma) 
\end{equation}
 
Substituting \eqn{MKe13} into \eqn{MKe12}, we obtain: 
\begin{equation}  \label{MKe14}
m ^{\downarrow p \cup q}(a) = 
\sum_{
\SP{
$b,\gamma $;\\
$ b \subseteq \Xi_p \times \Xi_q,$\\
$\gamma  \subseteq  \Xi_p, $\\
$a =  b  \cap \gamma  ^{\uparrow p \cup q}  $
}
}
 m'(b) \cdot  m   ^{\downarrow p}( \gamma  ) 
\end{equation}

But from definition of conditionality (\eqn{MKe1}) 
and the definition of belief function (see def.\ref{DEFm})
we know that:
\begin{equation}  \label{MKe15}
m ^{\downarrow p \cup q}(a) = 
\sum_{
\SP{
$b,\gamma $;\\
$ b \subseteq \Xi_p \times \Xi_q,$\\
$\gamma  \subseteq  \Xi_p, $\\
$ a =  b  \cap \gamma  ^{\uparrow p \cup q}  $
}
}
m ^{\downarrow p \cup q | p} (b) \cdot  
m   ^{\downarrow p}( \gamma  ) 
\end{equation}

Hence, by comparison of \eqn{MKe14} and \eqn{MKe15} we conclude that 
 $m'$ must be the mass function of a conditional belief function
$Bel ^{\downarrow p \cup q | p}$ so the claim of the theorem  is 
proven. 
\EndBeweis

The above theorem has an existential form: if the compressible independence
of conditional belief on a variable is given then there exists the compression
similar to Lemma 3.1. claim 7 of \cite{Shenoy:94} in which valuation
identity is replaced by group identity even for
normal valuations.

Let us notice that under the conditions of the above theorem 
(combining \eqn{MKe1} and \eqn{MKe3})

 \begin{equation}
 Bel = 
Bel ^{|p \cup r}\oplus Bel ^{\downarrow p \cup r } =
  Bel ^{\downarrow p \cup q | p} 
\oplus Bel ^{\downarrow p \cup r }
\end{equation}

\noindent
and hence for {\bf any } $Bel ^{\downarrow p \cup r | p}$
\begin{equation}
Bel =  
  Bel ^{\downarrow p \cup q | p}
\oplus Bel ^{\downarrow p}
\oplus Bel ^{\downarrow p \cup r | p }
\end{equation}

and therefore
\begin{equation}
Bel =  
  Bel ^{\downarrow p \cup q} 
\oplus Bel ^{\downarrow p \cup r | p }
\end{equation}

This means that whenever the conditional 
$Bel ^{p \cup q \cup r|p \cup r}$ 
is compressibly independent of $r$, 
then there exists a 
conditional 
$Bel ^{p \cup q \cup r|p \cup q}$ 
compressibly independent of $q$.
But this fact  combined with the previous theorem results in:

\Bem{
\begin{th} \label{MKt2}
Let $p,q,r$ be pairwise disjoint sets of variables and 
let $V =p \cup q \cup r$ and let $Bel$ be defined 
over V.
Furthermore let $Bel ^{|p \cup r}$ be a conditional Belief conditioned 
on variables $p\cup r$. Let this conditional distribution be 
compressibly independent of 
$r$. 
Then the vacuous extension onto $V$ of any 
 $Bel ^{\downarrow p \cup q | p}$  being a conditional belief of 
projected belief conditioned on the variable $p$ 
is a conditional belief function of $V$  conditioned 
on variables $p,r$. Hence for every $A\subseteq \Xi$
$$ \frac{ Q(A) }
{ Q ^{\downarrow p \cup r}(A ^{\downarrow p \cup r} ) }
=  \frac { Q ^{\downarrow p \cup q}(A ^{\downarrow p \cup q} ) }
{ Q ^{\downarrow p}(A ^{\downarrow p} ) }
$$
\end{th}
}

\begin{th} \label{MKt2}
Let $p,q,r$ be pairwise disjoint sets of variables.
Let $V =p \cup q \cup r $ and let $Bel$ be defined 
 over V.
Furthermore let $Bel ^{|p \cup r}$ be an (anti)conditional Belief
conditioned on variables $p\cup r$. Let this conditional distribution be
 compressibly 
 independent of 
$r$. Let $Bel ^{\downarrow p \cup q}$ be the projection of $Bel$ onto 
the subspace spanned by $p\cup q$. Then, {\em for every} 
 $Bel ^{\downarrow p \cup q | p}$  being a conditional belief of that 
projected belief conditioned on the variables $p$
its  vacuous  extension,   
$(Bel ^{\downarrow p \cup q | p}) ^{\uparrow V}$
%
is  an (anti)conditional belief function of $Bel$
conditioned on variables $p\cup r$.
\end{th}

We can easily check that Shenoy's notion of conditionality implies
existence of conditional compressibly independent of a variable. 

\begin{th} \label{MKt3}
Let $p.q.r$ be three disjoint sets of variables.
Let $Bel$ be a belief function over space spanned by variables $p\cup q
\cup r$.
$q,r$ are Shenoy-independent given $p$ iff there exist $Bel ^{\downarrow
p\cup r|p}$  and  $Bel ^{
|p\cup q}$ such that  ($Bel ^{\downarrow
p\cup r|p })^{\uparrow p\cup q \cup r} = Bel ^{|p\cup q}$ (that is there
exists conditional on $p\cup q$ compressibly independent of $q$) 
\end{th}
\AnfBeweis
By definition, 
$q,r$ are Shenoy-independent given $p$ iff there exist (pseudo-)
 belief functions  $Bel_2$ over space $p\cup q$
 $Bel_3$ over space $p \cup r$ such that     
$Bel = Bel_2 \oplus Bel_3$.
So the if-part is obvious given definition \ref{MKd1}.
 But $Bel = Bel_2 \oplus Bel_3$ implies also that
$Bel = (Bel_2  ^{|p} \oplus Bel_2  ^{\downarrow p}) \oplus Bel_3$.
Hence 
$Bel = Bel_2  ^{|p} \oplus (Bel_2  ^{\downarrow p} \oplus Bel_3)$.
We choose the one $Bel_2  ^{|p}$ for which
after division but before normalization 
 $m_2
^{|p}(\Xi_p\times\Xi_q)=1$ 
and otherwise  $m_2 ^{|p}(A)=0$ whenever 
 $Q_2 ^{\downarrow p}(A ^{\downarrow p})=0$ 
 (such one always exists). Clearly 
then $(Bel_2  ^{|p})^{\downarrow p}$ is the vacuous belief function.
But $Bel ^{\downarrow p\cup q} =  Bel_2  ^{\downarrow p} \oplus Bel_3$.
Hence  $Bel_2  ^{|p}$ is in fact a $Bel ^{| p,q}$ and therefore there
exists a  conditional compressibly independent of $q$.
\EndBeweis

Let us now have a look at the intersection property required by Pearl and Paz
\cite{Pearl:87} for graphoidal structures. We insist here that we will work
with more general DS valuations than Shenoy \cite{Shenoy:94} did, that is
we explore the space of normal DS valuations. First, however, let us look more
closely at the very notion of independence.\\
We associate usually independence/dependence  with the freedom/slavery
concepts. As (next to) great philosophers suggest and as life confirms,
usually absolute freedom and absolute slavery coincide. Speaking more 
seriously,
 there are cases in probability calculus where dependence and independence
cannot be distinguished. Two variables X,Y are usually said to be
statistically independent when, whatever the value of X and Y, always: $P(X
\& Y)=P(X)\cdot P(Y)$. Now let have $P(X=x)=1$ and $P(X=\lnot x)=0$,
$P(Y=y)=p$ and $P(Y=\lnot y)=1-p$, Clearly, X and Y are in that sense
statistically independent - but they are possibly functionally dependent (We
can establish a function $f:Y\rightarrow X$ fitting the joint distribution of
X and Y)!  A still worse case is when three (binary) variables, X,Y,Z, are
connected by XOR relationship: X xor Y =Z, with X,Y being uniformly
distributed ($P(X=x \& Y=y) = P(X=\lnot x \& Y=\lnot y)  = P(X= x \&
Y=\lnot y) =  P(X=\lnot x \& Y= y) = 0.25$). This would suggest that X,Y are
independent. But also X,Z are then "independent" as well as Y,Z. Still another
peculiarity occurs when X=Y and Y=Z. Then X,Y are conditionally independent
given Z,  X,Z are conditionally independent
given Y,  Z,Y are conditionally independent
given X. The latter case is, by the way, a justification why Shenoy did not
allow for a general normal valuation when considering intersection axiom of
graphoids. If we want to observe intrinsic independence, we need to see
diversity of behaviors. Otherwise we do not know whether no change in one's
behavior is a response or a selfishness of a variable. Therefore we need a
notion of diversity.

\begin{df}
A (proper or pseudo) belief function Bel defined over the space $\Xi$ spanned
by the set of variables $V$ is said to be diverse
(with respect to $V$)
 iff for every $\xi \in \Xi$ we have 
$Q(\xi) \ne 0$ (that is commonality of singleton sets is non-zero). 
\end{df}

Notice that the property of diversity is retained for both proper and 
pseudo belief functions for operations of vacuous extension and 
anticonditioning and combination of belief functions via Dempster rule of
combination, but it
is retained only for proper belief functions for operation of marginalization.

Under the conditions of this definition we say that 

\begin{df}
Let $p.q.r$ be three disjoint sets of variables.
Let $Bel$ be a belief function over space spanned by variables $p\cup q
\cup r$.
$q,r$ are intrinsically independent given $p$ iff
Bel is diverse (with respect to to the variables  $p\cup q
\cup r$) and 
 there exist $Bel
^{\downarrow p\cup r|p}$  and  $Bel ^{
|p\cup q}$ such that  ($Bel ^{\downarrow
p\cup r|p })^{\uparrow p\cup q \cup r} = Bel ^{|p\cup q}$ (that is there
exists conditional on $p\cup q$ compressibly independent of $q$) 
\end{df}

It is an easy task to check
exploiting results of Shenoy \cite{Shenoy:94} - 
 that intrinsic independence relation 
fulfills the graphoidal requirements of symmetry, decomposition,
weak union and contraction for proper belief functions, as operations of
marginalization and anticonditioning preserve the property of diversity.
 
 The last graphoidal property, intersection property, is proved below as
\begin{th} \label{MKt4}
Let $p,q,r,s$ be pairwise disjoint sets of variables.     
Let Bel be a proper belief function defined over the set of variables
$V=p\cup q \cup r \cup s$.
If $q$ and $s$ are intrinsically independent given $p\cup r$ and 
   $r$ and $s$ are intrinsically independent given $p\cup q$ then also 
 $q \cup r$ and $s$ are intrinsically independent given $p$.
\end{th}
\AnfBeweis
 Let us first notice that if a (pseudo-) belief function $Bel_1$
defined over the space spanned by variables $p\cup q$ ($p$ and $q$ disjoint)
is defined in
such a way that $Bel_1 = (Bel_1 ^{\downarrow p})^{\uparrow p\cup q}$ then for
 every
subset A of the discourse space $\Xi_p\times\Xi_q$ 
$Q_1(A)=Q_1((A ^{\downarrow p})^{\uparrow p\cup q})$ holds.\\
\begin{equation} \label{MKe20a}
Bel_1 = (Bel_1 ^{\downarrow p})^{\uparrow p\cup q}
\rightarrow 
Q_1(A)=Q_1((A ^{\downarrow p})^{\uparrow p\cup q})
\end{equation}
Now let us consider a function $Bel$ defined over  space  spanned 
by variables
$p,q,r,s$, where independence conditions  hold as required by the premise of
the theorem. Then definition \ref{MKd1} and theorem \ref{MKt3} imply
\begin{equation}
{Bel}^{\downarrow p\cup q\cup r} 
\oplus {Bel}^{\downarrow p\cup q\cup s | p\cup q}  =
Bel = 
{Bel}^{\downarrow p\cup q\cup r} 
\oplus {Bel}^{\downarrow p\cup r\cup s | p\cup r}  
\end{equation}

Let $V=p\cup q \cup r \cup s$ and let Bel be a function defined over the space
spanned by $V$. \\
Let us consider subsequently only unnormalized conditional Q's (commonality
functions, def. \ref{DEFm}), that is ones obtained by division: 
$Q ^{|p}(A)=\frac{Q (A)}{Q ^{\downarrow p}(A)}$. Unnormalized conditional Q's
differ from normalized ones only by a constant factor independent of the
function's argument.\\
Let us consider two sets $A_1,A_2 \subseteq \Xi_p\times\Xi_q\times\Xi_r
\times\Xi_s$  
such that $A_1^{\downarrow p\cup r\cup s} 
= A_2^{\downarrow p\cup r\cup s}$ 
and with $Q ^{\downarrow p\cup q\cup r\uparrow V}(A_1)>0$ and 
 $Q ^{\downarrow p\cup q\cup r\uparrow V}(A_2)>0$.
Then we have 
\begin{equation}
{Q}^{\downarrow p\cup q\cup r \uparrow V}(A_i)
\cdot  {Q}^{\downarrow p\cup q\cup s | p\cup q  \uparrow V}
(A_i) 
  =
{Q}^{\downarrow p\cup q\cup r \uparrow V}(A_i)
\cdot  {Q}^{\downarrow p\cup r\cup s | p\cup r\uparrow V} 
(A_i)
\end{equation}
for i=1,2, which is easily simplified to 
\begin{equation} \label{MKe22}
 {Q}^{\downarrow p\cup q\cup s | p\cup q\uparrow V}(A_i)
  =
  {Q}^{\downarrow p\cup r\cup s | p\cup r\uparrow V}(A_i) 
\end{equation}
As stated previously (\eqn{MKe20a}), however 
\begin{equation} \label{MKe23}
Q ^{\downarrow p\cup r\cup s | p\cup r\uparrow V}
(A_i)=
Q   ^{\downarrow p\cup r\cup s | p\cup r\uparrow V}
(A_i ^{\downarrow p\cup r\cup s\uparrow V})
\end{equation}

But due to the assumption that 
 $A_1^{\downarrow p\cup r\cup s} 
= A_2^{\downarrow p\cup r\cup s}$.
we get from \eqn{MKe23}
\begin{equation} \label{MKe24}
Q ^{\downarrow p\cup r\cup s | p\cup r\uparrow V}
(A_1)=
Q ^{\downarrow p\cup r\cup s | p\cup r\uparrow V}
(A_2)
\end{equation}
and by substituting \eqn{MKe24} into \eqn{MKe22} 
\begin{equation} \label{MKe25}
Q ^{\downarrow p\cup q\cup s | p\cup q\uparrow V}
(A_1)=
Q ^{\downarrow p\cup q\cup s | p\cup q\uparrow V}
(A_2)
\end{equation}

Let us consider two sets $A_1,A_2 \subseteq \Xi_p\times\Xi_q\times\Xi_r
\times\Xi_s$  
such that $A_1^{\downarrow p\cup q\cup s} 
= A_2^{\downarrow p\cup q\cup s}$ 
and with $Q ^{\downarrow p\cup q\cup r\uparrow V}(A_1)$ and 
 $Q ^{\downarrow p\cup q\cup r\uparrow V}(A_2)>0$.
Then we have (by similar argument)
\begin{equation} \label{MKe25b}
Q ^{\downarrow p\cup r\cup s | p\cup r\uparrow V}
(A_1)=
Q ^{\downarrow p\cup r\cup s | p\cup r\uparrow V}
(A_2)
\end{equation}

Now we can say that if for two sets  $A_1,A_2 \subseteq
\Xi_p\times\Xi_q\times\Xi_r \times\Xi_s$  
with $Q ^{\downarrow p\cup q\cup r\uparrow V}(A_1)>0$ and 
 $Q ^{\downarrow p\cup q\cup r\uparrow V}(A_2)>0$.
we have always 
\begin{equation} \nonumber     
Q ^{\downarrow p\cup q\cup s | p\cup q\uparrow V}
(A_1)=
Q ^{\downarrow p\cup q\cup s | p\cup q\uparrow V}
(A_2)
\end{equation}
whenever we can establish a path $B_1=A_1,B_2,...,B_n=A_2$ such that
for all i=1,...,n  $Q ^{\downarrow p\cup q\cup r\uparrow V}(B_i)>0$ and 
for all i=1,...,n-1
either  $B_i ^{\downarrow p\cup r\cup s} 
= B_{i+1}^{\downarrow p\cup r\cup s}$.%
or   $B_i ^{\downarrow p\cup q\cup s} 
= B_{i+1}^{\downarrow p\cup q\cup s}$.%

Let us now consider those sets  $A \subseteq
\Xi_p\times\Xi_q\times\Xi_r \times\Xi_s$  
with $Q ^{\downarrow p\cup q\cup r\uparrow V}(A)=0$
Then                      $Q ^{| p\cup q\cup r}(A)$ 
may be assigned any value. However, to prove the claim of the theorem, we need
to assign such a value that 
\begin{equation} \nonumber 
Q ^{\downarrow p\cup q\cup s | p\cup q\uparrow V}
(A)=
Q ^{\downarrow p\cup q\cup s | p\cup q\uparrow V}
(A')
\end{equation}
for every $A' \subseteq
\Xi_p\times\Xi_q\times\Xi_r \times\Xi_s$  
with $Q ^{\downarrow p\cup q\cup r\uparrow V}(A') \ge 0$.
and with 
 $A   ^{\downarrow p\cup r\cup s} 
= {A'}     ^{\downarrow p\cup r\cup s}$.%
This means, that we have to meet the requirement that for every path 
 $B_1=A_1,B_2,...,B_n=A_2$ such that
for all i=1,...,n-1
either  $B_i ^{\downarrow p\cup r\cup s} 
= B_{i+1}^{\downarrow p\cup r\cup s}$.%
with 
$Q ^{\downarrow p\cup q\cup s | p\cup q\uparrow V}
(B_i)=
Q ^{\downarrow p\cup q\cup s | p\cup q\uparrow V}
(B_{i+1})
$
or   $B_i ^{\downarrow p\cup q\cup s} 
= B_{i+1}^{\downarrow p\cup q\cup s}$.%
with 
$Q ^{\downarrow p\cup r\cup s | p\cup r\uparrow V}
(B_i)=
Q ^{\downarrow p\cup q\cup s | p\cup q\uparrow V}
(B_{i+1})
$.
As sets $B_i$ with  $Q ^{\downarrow p\cup q\cup r\uparrow V}(B_i)=0$ cause no
trouble (their conditional Q-values may be manipulated), the only difficulty
may stem from $B_i$s with  $Q ^{\downarrow p\cup q\cup r\uparrow V}(B_i)>0$
so that for all paths  \{$B_i$\} we need to have a special path 
\{$B'_i$\} such
that all of $B_i$s with  $Q ^{\downarrow p\cup q\cup r\uparrow V}(B_i)>0$ from
any possible path belong to a subpath $B'_j,B'_{j+1},...,B'_{j+m}$ with 
 $Q ^{\downarrow p\cup q\cup r\uparrow V}(B'_{j+k})>0$ for every k=0,...,m.\\

 But the existence of B'-path is a straight forward
consequence of the diversity assumption.

Hence we can always construct such a conditional that 
\begin{equation} 
Q ^{\downarrow p\cup q\cup s | p\cup q\uparrow V}
(A_1)=
Q ^{\downarrow p\cup q\cup s | p\cup q\uparrow V}
(A_2)
\end{equation}
for every pair of sets $A_1,A_2$ such that  $A_1^{\downarrow p\cup
r\cup s} = A_2^{\downarrow p\cup r\cup s}$, and especially
if $A_2 = 
(A_2^{\downarrow p\cup r \cup s})^{\uparrow  p\cup q\cup r \cup s}$.     But
the latter means  that  this conditional 
$Bel ^{\downarrow p\cup q\cup s | p\cup q\uparrow V}$ is
compressibly independent of $q$, so that in fact there exists 
a conditional that 
\begin{equation}  \label{MKe26}
Bel ^{\downarrow p\cup q\cup s | p\cup q\uparrow V}=
Bel ^{\downarrow p\cup s | p\uparrow V}
\end{equation}

This implies again that:
\begin{equation}
Bel = Bel ^{\downarrow p\cup s | p} \oplus 
{Bel}^{\downarrow p\cup q\cup r}
\end{equation}
which implies via theorem \ref{MKt3} that  $q \cup r$ and $s$ are independent
given $p$.
\EndBeweis

\section{Discussion}

Two     approaches of structuring (factorization, decomposition) of
Dempster-Shafer joint belief functions from literature,
of Cano et al. \cite{Cano:93} and of Shenoy \cite{Shenoy:94}, 
 have been   
reviewed
 with
special emphasis on their capability to capture and exploit  independence 
 for purposes of factorization in terms of graphoidal structure. 
 It has been demonstrated that in Cano et al. \cite{Cano:93} framework
some belief functions factorable graphoidally in the sense of Shenoy
\cite{Shenoy:94}
 cannot 
be factored and hence do not correspond to conditional independence in the
sense of Cano et al. \cite{Cano:93}. On the other hand,
though conditional independence is defined in a much broader sense in Shenoy's
paper \cite{Shenoy:94}, 
 Shenoy demonstrates
that
his notion of independence is a graphoidal relation, but only for
positive normal valuations. This actually means that probabilistic belief
functions, as not  possessing  positive  normal  valuations,  are 
actually
excluded from consideration. Shenoy's and Cano et al.'s notions of
(graphoidal) independence  coincide for positive normal valuations whenever
respective Cano et al.s (a priori) conditional belief function exists.

The exclusion of probabilistic belief functions from graphoidal structuring is
remarkable because of the general claim that DS belief functions constitute a
generalization of probability distributions. For
VBS consisting of 
 probability distributions as
such, Shenoy's notion of positive valuation has been identified as requirement
of a probability distribution without null values in any cell. Why should then
probabilistic belief functions
within VBS of DS belief functions
 with non-zero masses at every singleton not
fulfill requirements of graphoidal independence ? 

Widening of the class of VBS of DS belief functions in such a way 
as to make probabilistic belief functions
 with non-zero masses at every singleton 
fulfill requirements of graphoidal independence was one of goals of this
investigation. To achieve this goal,  
this paper verifies the notion of independence in that it requires that the
Q-function (commonality function) is not null for singleton sets. In theorem
\ref{MKt4} it has been demonstrated that such a notion of independence
fulfills the requirement of intersection, the only one property of Shenoy's
notion of independence for which positive normal valuation is required. This
new notion of independence   covers clearly the  Shenoy's  notion 
of positive 
normal independence   as a special cases, because in proper belief functions
$Q(\Xi)>0$ implies  $Q(A)>0$ for every $A \subseteq\Xi$, including all
singletons. Also, probabilistic
independence in probabilistic 
 belief functions
 with non-zero masses at every singleton qualifies as a special case
of the new notion of (intrinsic) independence. 

As a pre-requisite for this result,
notion of conditionality as such has been revisited. Instead of Cano's
a priori conditional belief functions and Shenoy's (normal) conditional
belief functions  a broader notion of (anti)conditional belief functions has
been introduced. Both Shenoy's and Cano's conditionals can be treated as
special cases of conditionals introduced in definition \ref{MKd1}. 
Compared to Cano et al's notion of (a priori) conditionality, we must state
that whenever Cano conditional exists, our exists, but not vice versa.\\
One difference to Shenoy's
approach is important: we do not require that there exists a unique
conditional for a given belief function and the set of conditioning
variables (we do not require positive valuations). 
 Under these circumstances, if graphoidal properties are to be
demonstrated, a shift from equality relations to existential equality
relations has to be made. In this spirit, it has been demonstrated that the
graphoidal property of intersection is fulfilled for conditional independence
relationship not only for Shenoy's positive normal but also for Shenoy's
normal valuations with positive Q's on singleton sets.  Hence one can
conclude that a much broader class
of conditional factorizations of belief functions has  graphoidal 
properties
 than those with Cano's specific a-priori conditionals.\\

Widening the notion of conditionality from a single function to a family of
functions has several consequences for general normal valuations. In
probability calculus, if variables X,Y are independent given Z, then
we understand that $P(X|Z,Y)=P(X|Z)$ that is conditional of X given Z,Y can
be derived from X,Z alone. Given Shenoy's
notion of conditionality, an equation like this is not valid for DST, as 
 $r \indep s | v$ is equivalent to
 $\gamma(r|s \cup v)=\gamma(r|v) \oplus \tau_{\gamma(s\cup v)}$
that is knowledge of r,v alone ($\gamma(r|v)$) is insufficient to construct 
 $\gamma(r|s \cup v)$ (because member identity $\tau_{\gamma(s\cup v)}$
of $\gamma(s\cup v)$ is also required). 
However, under theorem \ref{MKt1}, in the class of
conditionals given by definition \ref{MKd1} this is possible - if variables r,
s are independent given v, then a (and via theorem \ref{MKt2} every) $Bel
^{\downarrow r\cup v |v}$ is a legitimate  $Bel ^{|v\cup s}$. \\
Furthermore, in probability theory conditional probability may be viewed as a
kind of generalization of knowledge, "freeing" the experience from the
particular distribution of the conditioning variable. Invariance of the
conditional distribution over various samples indicates detection of intrinsic
relationship. Given Shenoy's conditioning of belief functions, even if we have
an intrinsic relationship among variables, we will get different conditional
belief functions for different "samples" of joint belief distribution. On the
other hand, the definition \ref{MKd1} of conditionality ensures that in such
cases the various samples will share (at least one) common (anti)conditional.
(This is due to theorem \ref{MKt1} as it corresponds to
compressible independence of a variable levels of which generate these sample
belief functions.)\\
We cannot overlook that, under validity of theorems \ref{MKt1}-\ref{MKt4}, 
the model of decomposition of DS belief functions proposed in
\cite{Klopotek:93f} combines the merits
of both Cano et al. \cite{Cano:93} and of Shenoy/Shafer \cite{Shenoy:90}
approaches to decomposition of DS belief functions  
as on the one hand 
no simpler factorization (in terms of number of variables in hypernodes) into
a hypertree
of Shenoy/Shafer (hence for propagation of uncertainty using their method)
exists 
 than  one consisting of conditional factors (paralleling bayesian networks)
proposed in \cite{Klopotek:93f}; and on the other hand   
the decomposition proposed  in \cite{Klopotek:93f} captures (conditional and
unconditional) independence among variables 
for a much broader class of belief distributions than Cano et al. framework
does.

Some words must be said about disadvantages of the intrinsic conditional
independence. While Shenoy's positive normal independence requires only to
check for presence of a single focal point (the universe focal point), the
intrinsic independence requires checking every singleton set of the universe
(which may not necessarily be a focal point of the distribution). 
The question may be formulated whether one could change Shenoy's normal 
valuation to positive normal valuation simply by adding a focal point 
for  the universe set. This question seems to have the answer NO as then e.g.
a probabilistic belief distribution with two 
unconditionally
independent variables, each having
domain with cardinality three or more would then  turn to a distribution in
dependent variables (unless one adds some other focal points). 

Further research concerning the class of valuations possessing notion of
conditional independence and fulfilling graphoidal axioms seems to be
necessary. In particular we can ask, whether one, or two or more Q-values of
singletons equal zero will harm the graphoidal properties. One should also ask
what
can be concluded about graphoidal properties if we are unable to investigate
all focal points of the whole  distribution, but only of its projections onto
subsets of the set of variables containing up to, say, k variables ?
Currently we can say that if we were able to construct a belief network
of the type defined in \cite{Klopotek:93f}, and are able to verify that each
factor in this belief network factorization fulfills the requirement of
diversity, then the combined distribution of all factors will do.  However,
we cannot ensure (by investigating subsets of variables with cardinality up to
k only) that the combined distribution is in fact a proper belief function -
we can only check that this is a pseudo-belief distribution. This means that 
projections of the combined distribution may fail to be diverse.

\section{Conclusions}

\begin{enumerate}
\item 
A new notion of conditionals (anticonditionals) has been introduced for
Dempster-Shafer belief functions. It is characterized by the fact that in
general many belief functions can be considered as a conditional belief
function of a given belief function.
\item 
Both Shenoy's \cite{Shenoy:94} and Cano's \cite{Cano:93} conditionals
can be
treated as special cases of conditionals introduced in this paper.
\item 
In the new definition of conditionality,  if variables $r$,
$s$ are independent given $v$, then every conditioning of 
the belief function marginalized onto $v \cup r$  on $v$
is a legitimate conditional for the original belief function
conditioned on  $v \cup s$. This property is  not valid for Shenoy's notion
of conditioning.
\item The notion of compressible independence of a belief distribution from a
variable has been introduced in that a belief function $Bel$ defined over the
space spanned by the set of variables $V$ is compressibly independent of a
subset $p$ of $V$ iff $Bel ^{\downarrow V-p \uparrow V}=Bel$.
\item 
A new notion of conditional independence (intrinsic independence)
for proper belief functions 
 has been
introduced characterized by the fact that 
beside existence of a compressibly independent conditional also 
the commonality function shall take 
 non-zero values at all singleton sets.
\item  
For the DS belief functions, intrinsic independence relation  fulfills the
graphoidal axioms of \cite{Pearl:87}.
\item  
This
new notion of  intrinsic (conditional) independence   
generalizes  the
Shenoy's notion of positive 
normal independence  with the latter as its special case
\item  
Also, probabilistic
independence in probabilistic 
 belief functions
 with non-zero masses at every singleton qualifies as a special case
of the new notion of (intrinsic) independence - hence having graphoidal
properties within DS belief function framework. Probabilistic belief functions
are not positive normal valuations in the sense of Shenoy \cite{Shenoy:94},
hence were not proven to have this property within Shenoy's VBS framework
(though at the same time probability distributions had this property within
Shenoy's VBS).
\end{enumerate}


\begin{thebibliography}{99}
%
\newcommand{\LitStelle}[2]{\bibitem{#1}}
\newcommand{\A}[2]{#2 #1}
\newcommand{\IN}{[in:]}
\newcommand{\ReadingsIn}{G. Shafer, J. Pearl eds: 
{\it Readings in Uncertain 
Reasoning}, (ISBN 1-55860-125-2, 
Morgan Kaufmann Publishers Inc., San Mateo, California, 1990)}
%
\LitStelle{Cano:93}{Cano et al., 1993}
\A{J.}{Cano}. \A{M.}{Delgado}, \A{S.}{Moral}: An axiomatic framework for
propagating 
uncertainty in directed acyclic networks,  {\it International Journal of 
Approximate Reasoning}. 1993:8, 253-280.
\LitStelle{Dawid:79}{Dawid, 1979}  
\A{A.P.}{Dawid}: Conditional independence in statistical theory (with
discussion), {\it J. Roy. Stat. Soc.}, Ser. B. 4(1), 1-31, 1979
\LitStelle{Hunter:91}{Hunter, 1991} 
\A{D.}{Hunter}: Graphoids and natural conditional functions, {\it Int. J.
Approx. Reas.}5(6),489-504, 1991
\LitStelle{Klopotek:93f}{Klopotek, 1994}  
\A{M.A.}{K{\l}opotek}: 
Beliefs in Markov Trees - From Local Computations to Local Valuation,
\IN  R. Trappl, Ed.: 
{\it Cybernetics and Systems Research}, 
%
World Scientific Publishers,  Vol.1.
pp. 351-358
\LitStelle{Pearl:87}{Pearl \& Paz, 1987} 
\A{J.}{Pearl},\A{A}{Paz}: Graphoids, graph-based logic for reasoning about
relevance relations, \IN {\it Advances in Artificial Intelligence-II},
B.D.Boulay, D.Hogg, L.Steele, Eds: North Holland, Amsterdam, 357-363, 1987.
 \LitStelle{Pearl:88}{Pearl, 1988}
\A{J.}{Pearl}:  {\it Probabilistic Reasoning in Intelligent Systems:Networks
of Plausible Inference}, Morgan Kaufmann, San Mateo CA, 1988
\LitStelle{Shafer:90b}{Shafer \& Srivastava, 1990b} 
\A{G.}{Shafer} , \A{R.}{Srivastava}: 
The Bayesian and Belief-Function Formalisms. A 
General Prospective for Auditing, in: \ReadingsIn, 482-521.
%
\LitStelle{Shenoy:90}{Shenoy \& Shafer, 1990}  
 \A{P.P.}{Shenoy}, \A{G.}{Shafer}: 
 Axioms for probability and belief-function propagation, \IN
 R.D. Shachter, T.S. Levit, L.N. Kanal, J.F. Lemmer eds:
{\it Uncertainty in Artificial Intelligence} 4,
(Elsevier Science Publishers B.V. (North Holland), (1990), 169-198.
%
\LitStelle{Shenoy:94}{Shenoy, 1994}  
 \A{P.P.}{Shenoy}:
Conditional independence in valuation based systems,
{\it International Journal of Approximate Reasoning}, Vol. 109, No. 3,
May 1994,
%
\LitStelle{Spohn:88}{Spohn, 1988} 
\A{W.}{Spohn}: Ordinal conditional functions: a dynamic theory of epistemic
states, \IN {\it Causation in Decision, Belief Exchange and Statistics}
(W.L.Harper, B.Skyrms Eds.) D.Reidl, Dordrecht, Holland,2,105-134, 1988
\LitStelle{Studeny:89}{Studeny, 1989} 
\A{M.}{Studeny}: Multiinformation and the problem of characterization of
conditional independence relation, {\it Problems of Control and Information
Theory}, 18(1),3-16,1989
%
\end{thebibliography}
\end{document}